\documentclass[submission,copyright,creativecommons]{eptcs}

\providecommand{\event}{Workshop of ThEdu'22}
\usepackage[english]{babel}
\usepackage[utf8]{inputenc}
\usepackage[T1]{fontenc}
\usepackage{amsthm} % proofs
\usepackage{tikz}
\usepackage{longtable}

\usetikzlibrary{arrows.meta,
                chains,
                positioning,
                quotes,
                shapes.geometric}
\usetikzlibrary{decorations.markings}
\usetikzlibrary{calc}
\usetikzlibrary{arrows}

\tikzstyle{line} = [draw,-Triangle] %draw, -latex']

\makeatletter
\tikzset{FlowChart/.style={
suspend join/.code = {\def\tikz@after@path{}},
     base/.style = {draw, fill=##1,
                    minimum height=9mm, text width=36mm,
                    align=center,
                    on chain, join=by arr
                   },
startstop/.style = {base=red!30},
  process/.style = {base=orange!30, rounded corners},
 decision/.style = {diamond, aspect=1.3, inner xsep=0pt,
                    draw, fill=green!30, align=center,  % since not use "base"
                    on chain, join=by arr},             % since not use "base"
       io/.style = {base=blue!30, trapezium, trapezium stretches body,
                    trapezium left angle=70, trapezium right angle=110},
      arr/.style = {semithick,-Triangle},
every edge quotes/.style = {auto, font=\footnotesize}
       }   }
\makeatletter

\newcounter{conjecturas}

\usepackage{graphicx}
\usepackage{color}

\definecolor{lightblue}{RGB}{231,255,255}
\definecolor{lightred}{RGB}{255,224,224}
\definecolor{lightgreen}{RGB}{224,255,224}
\definecolor{lightyellow}{RGB}{255,255,224}
\definecolor{lightpurple}{RGB}{255,224,255}
\definecolor{darkerred}{RGB}{64,0,0}
\definecolor{darkred}{RGB}{128,0,0}
\definecolor{darkblue}{RGB}{0,0,128}
\definecolor{darkgreen}{RGB}{0,128,0}
\definecolor{darkpurple}{RGB}{128,0,128}
\definecolor{black}{RGB}{0,0,0}

\usepackage{listings}

\newtheorem{theorem}{Theorem}
\newtheorem*{theorem*}{Theorem}

\title{A Rule-based Theorem Prover: an Introduction to Proofs in
  Secondary Schools}

\author{
  Joana Teles\thanks{This work is partially supported by the Centre
    for Mathematics of the University of Coimbra - UIDB/00324/2020,
    funded by the Portuguese Government through FCT/MCTES.}
  \institute{
    CMUC / Department of Mathematics, \\
    University of Coimbra, Portugal}
  \email{jteles@mat.uc.pt}
  \and
  Vanda Santos
  \thanks{This work is funded  by national funds through the FCT -
    Foundation for Science and  Technology, I.P., within the scope of
    the project UIDB/00194/2020 and in the scope of the framework
    contract foreseen in the numbers 4, 5 and 6 of the article 23, of
    the Decree-Law 57/2016, of August 29, changed by Law 57/2017, of
    July 19. }
  \institute{
    CIDTFF,  University of Aveiro \\
    and CISUC, Portugal}
  \email{vandasantos@ua.pt}
  \and
  Pedro Quaresma\thanks{This work is funded by national funds through
    the FCT - Foundation for Science and Technology, I.P., within the
    scope of the project CISUC - UID/CEC/00326/2020 and by European
    Social Fund, through the Regional Operational Program Centro
    2020.}
  \institute{
    CISUC / Department of Mathematics, \\
    University of Coimbra, Portugal}
  \email{pedro@mat.uc.pt}
}

\def\authorrunning{J. Teles, V. Santos \& P. Quaresma}

\def\titlerunning{A Rule-based TP: an Introduction to Proofs in
  Secondary Schools}

\begin{document}

% definição da linguagem de programação
\lstdefinelanguage{GDDM}{
  extendedchars=true,
  inputencoding=latin1,
  morekeywords={fof,axiom,para,parallelogram,quadrilateral,eqangle,simtri,cong,contri,rightangle,rectangle,perp,altintangles,coll,include,conjecture},
  basicstyle=\footnotesize
}

% definição da linguagem de programação
\lstdefinelanguage{Algoritmo}{
  extendedchars=true,
  inputencoding=latin1,
  morekeywords={WHILE,DO,ENDWHILE,IF,THEN,ELSE,ENDIF},
  basicstyle=\footnotesize
}

% % definição da linguagem de programação
% \lstdefinelanguage{FOF}{
%   extendedchars=true,
%   inputencoding=latin1,
%   morekeywords={include,fof,midp,para,conjecture},
%   basicstyle=\footnotesize
% }

\maketitle

\begin{abstract}
  The introduction of automated deduction systems in secondary schools
  faces several bottlenecks. Beyond the problems related with the
  curricula and the teachers, the dissonance between the outcomes of
  the geometry automated theorem provers and the normal practice of
  conjecturing and proving in schools is a major
  barrier to a wider use of such tools in an educational environment.

  Since the early implementations of geometry automated theorem
  provers, applications of artificial intelligence methods, synthetic
  provers based on inference rules and using forward chaining
  reasoning are considered to be best suited for education
  proposes.

  Choosing an appropriate set of rules and an automated method that
  can use those rules is a major challenge. We discuss one such rule
  set and its implementation using the geometry deductive databases
  method (GDDM). The approach is tested using some chosen geometric
  conjectures that could be the goal of a 7th year class ($\approx$12-year-old
  students). A lesson plan is presented, its
  goal is the introduction of formal demonstration of proving geometric
  theorems, trying to motivate students to that goal.
 \end{abstract}

\section{Introduction}
\label{sec:introduction}

The introduction of automated deduction systems in secondary schools
faces several bottlenecks. The absence of the subject geometry itself, of rigorous
mathematical demonstrations, not to mention formal proofs, in many of
the national curricula, the lack of knowledge (and/or training) by the
teachers about the subject~\cite{Santos2021}, the dissonance
between the outcomes of the available Geometry Automated Theorem
Provers (GATP) and the normal practice of conjecturing and proving in the
secondary education~\cite{Quaresma2022a}, are the most important in
our opinion.

According to~\cite{mec2013}, the structuring of thinking stands out as
one of the major purposes for the teaching of mathematics. In order to
achieve it, teaching should be based on sequential learning, to build knowledge in the classroom, a
hierarchy of concepts with the systematic study of their properties,
fostering clear and precise arguments. These are the basic elements of
hypothetical-deductive reasoning, which is par excellence the
mathematical reasoning. Inductive reasoning also plays an important
role in mathematics since it allows the establishment of conjectures,
which can then be proved using deductive reasoning. All forms of
reasoning have had, and continue to have, their place in the
mathematics curriculum~\cite{Hanna2020}.

The new curriculum documents~\cite{dge2018, dge2021} also reinforce
the importance of promoting and mobilising computational thinking,
assuming the ability to analyse and define algorithms, allowing a
structuring of thinking and providing students with more tools to
solve problems and prove results.  The use of technological resources
is unavoidable and the learning of mathematics can benefit from their
use. This work unites all of those purposes.

The paper is organised as follows:
first, in Sec.~\ref{sec:gr}, some dynamic geometry tools and  geometry automated theorem provers will be highlighted. In Sec.~\ref{sec:gddm}, the algorithm of a rule-based geometry automated theorem prover and the  set of rules  that will be used in the examples will be presented. In Sec.~\ref{sec:somethirdclassroomgeometryproblems}, two examples and excerpts of the formal proofs done by a rule-based geometry automated theorem prover will be analysed. In Sec.~\ref{sec:lessonPlanoutline} an outline of a lesson plan to address one of the examples will be proposed. Finally, in Sec.~\ref{sec:conclusions} conclusions are drawn and future work will be discussed.

\section{Geometric Reasoning}
\label{sec:gr}

Learning geometry involves some cognitive
  complexity. According to~\cite{Duval} there are three cognitive
  processes involved in learning geometry: visualisation (relating to spatial representation),
  construction (using tools) and reasoning (in particular the
  discursive processes to broaden the knowledge processes, for
  demonstration and for interpretation). These processes can be
  performed separately, that is, the visualisation does not depend on
  the construction. Even if construction precedes visualisation,
  construction processes only depend on connections between
  mathematical properties and tool constraints. If visualisation is an intuitive
  aid useful to find a proof, it can, in some cases, be misleading. The
  validity of the proof rely solely on the corpus of propositions
  (definitions, axioms, theorems) that are available.

  As Duval points out ``these three kinds of cognitive processes are
  closely connected and their synergy is cognitively necessary for
  proficiency in geometry''~\cite[p. 38]{Duval}.  This work aims to
  make contributions to the complex question of how to get
  $\approx\!12$-year-old students to understand and see the connections
  between these processes, in a given content of their curriculum.

\subsection{Dynamic Geometry Tools}
\label{sec:dgt}
 Dynamic geometry systems (DGS) have stimulated
  investigations into students' conceptions of mathematical
  demonstrations. There are studies that prove that this contribution
  of the DGS is twofold. First, they provide environments in which
  students can experiment freely, easily verifying their intuitions
  and conjectures in the process of looking for patterns, general
  properties, etc.  Second, they provide non-traditional ways for
  students to learn and understand mathematical concepts and
  methods~\cite{Marrades2000}. One of the advantages of DGS consists
  of carrying out tasks, not only for exploring geometric situations,
  but also for investigating situations that the tool itself promotes
  when moving objects, thus providing valuable support, both for
  students and for teachers.  From the several DGS available to support
  geometry learning, we can highlight: Cabri~\cite{Laborde1990},
  C.a.R.~\cite{Grothmann2016}, Cinderella~\cite{Richter-Gebert1999},
  GeoGebra~\cite{Hohenwarter2002}, GeometerSketchpad~\cite{Jackiw2001} and JGEx~\cite{Ye2011}.

\subsection{Geometry Automated Theorem Provers}
\label{sec:gatp}

  Automated deduction in geometry has been, since
  1960s, an important field in the area of automated
  reasoning. Various methods and techniques have been studied and
  developed for automatically proving and discovering geometric
  theorems~\cite{Quaresma2022e}. Focusing in the DGS/GATP platforms,
  i.e., platforms that combine the DGS with one (or several) GATP(s)
  we can highlight (some have been mentioned already): Cinderella, with a
  \emph{randomised prover}; GCLC~\cite{Janicic2006c}, which include several provers (area
  method, Wu's method and Gröbner basis method)~\cite{Janicic2012a};
  GeoGebra, which include several algebraic
  provers~\cite{Kovacs2018,Kovacs2022} and JGEx which include several provers (area method, full-angle method, deductive databases method, Wu's method and Gröbner basis method).
  In the last of these systems, the JGEx (\emph{Java Geometry Expert}), three components
  can be identified: the dynamic geometry component, the automated
  deduction component, and the most distinctive component, the
  generation of visually dynamic presentation of proofs in plane
  geometry.

  The JGEx will be used in the following sections, specifically its GATP based in the deductive database method.

\section{A Rule-based Geometry Automated Theorem Prover }
\label{sec:gddm}

Since the early attempts, linked to artificial intelligence, synthetic
provers based on inference rules and using forward chaining reasoning
has been seen has a more suited approach for education. Using an
appropriated set of rules and using forward chaining they can more
easily mimic the expected behaviour of a student when developing a
proof~\cite{Chou2001,Quaresma2022e}.

The JGEx geometry deductive database method prover is an efficient implementation
of a synthetic method, based on a set of inference rules.  For a given
geometric configuration, the program can find its fix-point with
respect to a fixed set of geometric inference rules, in other words,
it can find all the properties of the geometric construction
(conjecture) that can be deduced using those rules~\cite{Chou2000}.

After finding the fix-point, if the conjecture is among all the
deduced facts, the conjecture is proved. A synthetic proof can then be
generated with a natural language and visual renderings. The method is not complete, if the
conjecture is not among the deduced facts it does not mean  that it is not a theorem, it means that a different method, a decision procedure (e.g. the area method), must be used.

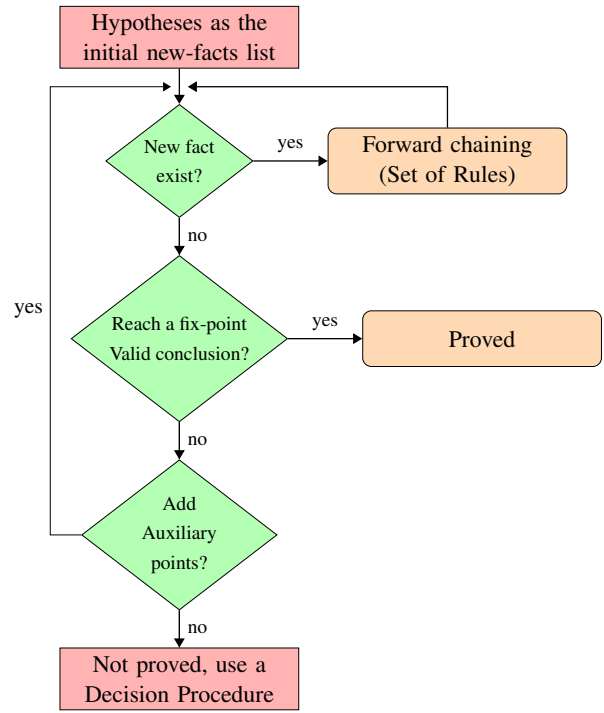
\begin{figure}[htbp!]
   \begin{center}
     \resizebox{0.5\textwidth}{!}{
       \begin{tikzpicture}[FlowChart,
         node distance = 6mm and 12mm,
         start chain = going below,
         ]
         \node (start) [startstop] {Hypotheses as the initial new-facts  list};
         \node (preDec1) [below of=start,  yshift=-0.2cm] {};
         \node (dec1)  [decision,below of=preDec1]    {\footnotesize New fact\\ \footnotesize exist?};
         \node (dec2) [decision]{\footnotesize Reach a  fix-point\\
           \footnotesize Valid conclusion?};
         % \node (rd)    [process]   {record\_data\\ (records, new\_record)};
         \node (dec3)  [decision]  {\footnotesize Add\\ \footnotesize
           Auxiliary\\
           \footnotesize points?};
         \node (stop)  [startstop] {Not proved, use a Decision Procedure};

         \node (proc1) [process, suspend join,right=of dec1] {Forward chaining\\ (Set of Rules)};
         \node (proc2) [process, suspend join,right=of dec2]   {Proved};

         \path   (dec1) edge ["no"]  (dec2)
         (dec2) edge ["no"]   (dec3)
         (dec3) edge ["no"]   (stop)
         (dec1) edge [arr, "yes"] (proc1)
         (dec2) edge [arr, "yes"]   (proc2);
         % (dec3) edge [arr, "yes"] (preDec1);
         \draw[arr] (proc1) |- (preDec1);
         %% \draw[arr] (dec3) |- (preDec1);
         \path[line] (dec3.west) -- ++(-0.5,0) coordinate(A) --
         node[midway]{\small yes\qquad\ } (A|-preDec1) -- (preDec1.west);
       \end{tikzpicture}}
   \end{center}
   \caption{Rule-based Geometry Automated Theorem Prover}
   \label{fig:rulebasedGATP}
 \end{figure}

The algorithm is simple, a data-based search strategy is used, a list
of ``new data'' is kept and for each new data the system searches the
rule set to find and apply the rules using
this data (see
Figure~\ref{fig:rulebasedGATP})~\cite{Chou2000}.

In the original implementation~\cite{Chou2000} and in a new
implementation being finished~\cite{Baeta2022}, a rule set based on the
full-angle method~\cite{Chou1996b} is used. Both implementation are open
source software, available in GitHub
servers.\footnote{JGEx:
  \url{https://github.com/yezheng1981/Java-Geometry-Expert}, OGP-GDDM:
  \url{https://github.com/opengeometryprover/OpenGeometryProver}}

A different set of rules can be used, keeping the efficient database
search strategy. In the following, the application of a set of
rules~\cite{Font2021,Font2018}, more adapted to the 7th year ($\approx$
12-year-old students) in the Portuguese
curricula~\cite{mec2013,dge2018,dge2021} is presented and its
application to the  demonstration of formal proof of some geometric theorems is described.

\subsection{A Set of Rules for the 7th Year}
\label{sec:thesetsofrules}

The set of rules presented here is contained in the set of rules
implemented in the tutorial system
\emph{QED-Tutrix}~\cite{Font2021,Font2018,Gagnon2017}.\footnote{Only the rules needed in the
  proofs were chosen.} Its formalisation in the \emph{TPTP
FOF-format} is presented here\footnote{\url{http://tptp.cs.miami.edu/TPTP/QuickGuide/Problems.html}}.
This set of rules is to be added to the set of rules used to implement the deductive database prover~\cite{Baeta2022,Chou2000}, notice that some of them are already taken (coincide) with rules presented in that set of rules. The \emph{FOF format} is chosen because it is the format used in the new implementation of the deductive database method
already mentioned above~\cite{Baeta2022}. The geometric predicates\footnote{para =
  parallel; perp = perpendicular; eqangle = equal-(full)angles; cong =
  congruent segments; simtri = similar triangles; contri = congruent
  triangles; coll = collinear} are those described in~\cite{Chou2000},
with some additions whose meaning should be clear.

\begin{itemize}
\item[\textbf{R1}] (definition of parallelogram) A quadrilateral
  $[ABCD]$ is a parallelogram iff $AB$ is parallel to $CD$ and $BC$ is
  parallel to $AD $. %(Ludovic 97)

\begin{lstlisting}[language=GDDM]
fof(ruleR1,axiom,(![A,B,C,D] :
           (para(A,B,D,C) & para(A,D,B,C) => parallelogram(A,B,C,D)) )).
fof(ruleR1a,axiom,(![A,B,C,D] :
           (parallelogram(A,B,C,D) => para(A,B,D,C)) )).
fof(ruleR1b,axiom,(![A,B,C,D] :
           (parallelogram(A,B,C,D) => para(A,D,B,C)) )).
\end{lstlisting}

\item[\textbf{R2}] If two lines are parallel, the alternate interior angles
  determined by a transversal are equal. %- (Ludovic 12)
  This is rule D40 of the deductive database method~\cite{Chou2000}.

\begin{lstlisting}[language=GDDM]
fof(ruleD40,axiom,(![A,B,C,D,P,Q] :
            (para(A,B,C,D) => eqangle(A,B,P,Q,C,D,P,Q)) )).
\end{lstlisting}

\item[\textbf{R3}] (a.s.a. criterion of equality of triangles). % - (Ludovic 38)
  This is rule D61 of the deductive database method~\cite{Chou2000}.

\begin{lstlisting}[language=GDDM]
fof(rulerD61,axiom,(![A,B,C,P,Q,R] :
             (simtri(A,B,C,P,Q,R) & cong(A,B,P,Q) => contri(A,B,C,P,Q,R)) )).
\end{lstlisting}

\item[\textbf{R4}] Given two equal triangles $[ABC]$ and $[DEF]$, the sides and
  the corresponding angles are equal $\overline{AB}=\overline{DE}$,
  $\overline{BC}=\overline{EF}$, $\overline{CA}=\overline{FD}$, and for
  the angles $B\widehat{A}C=E\widehat{D}F$, $C\widehat{B}A=F\widehat{E}D$ and $A\widehat{C}B=D\widehat{F}E$. % (ludovic 1)

\begin{lstlisting}[language=GDDM]
fof(ruleR4,axiom,(![A,B,C,P,Q,R] :
           (cong(A,B,P,Q) & cong(A,C,P,R) & cong(B,C,Q,R)
           => contri(A,B,C,P,Q,R)) )).
fof(ruleR4a,axiom,(![A,B,C,P,Q,R] :
           (contri(A,B,C,P,Q,R) => cong(A,B,P,Q) ) )).
fof(ruleR4b,axiom,(![A,B,C,P,Q,R] :
           (contri(A,B,C,P,Q,R) => cong(A,C,P,R)) )).
fof(ruleR4c,axiom,(![A,B,C,P,Q,R] :
           (contri(A,B,C,P,Q,R) => cong(B,C,Q,R)) )).
\end{lstlisting}

\item[\textbf{R5}] (definition of rectangle) A quadrilateral $[ABCD]$
  is a rectangle iff the interior angles are all right
  angles. % (Ludovic 105):

\begin{lstlisting}[language=GDDM]
fof(ruleR5,axiom,(![A,B,C,D] :
           (rightangle(D,A,A,B) & rightangle(A,B,B,C) &
            rightangle(B,C,C,D) & rightangle(C,D,D,A)
           => rectangle(A,B,C,D)) )).
fof(ruleR5a,axiom,(![A,B,C,D] :
           (rectangle(A,B,C,D) => rightangle(D,A,A,B)) )).
fof(ruleR5b,axiom,(![A,B,C,D] :
           (rectangle(A,B,C,D) => rightangle(A,B,B,C)) )).
fof(ruleR5c,axiom,(![A,B,C,D] :
           (rectangle(A,B,C,D) => rightangle(B,C,C,D)) )).
fof(ruleR5d,axiom,(![A,B,C,D] :
           (rectangle(A,B,C,D) => rightangle(C,D,D,A)) )).
fof(ruleR5e,axiom,(![A,B,C,D] :
           (rectangle(A,B,C,D) => para(A,B,D,C)) )).
fof(ruleR5f,axiom,(![A,B,C,D] :
           (rectangle(A,B,C,D) => para(A,D,B,C)) )).
\end{lstlisting}

\item[\textbf{R6}] If the angle $ABC$ is a right angle then the lines $AB$ and $BC$ are
  perpendicular. % (Ludovic 28)

\begin{lstlisting}[language=GDDM]
fof(ruleR6,axiom,(![A,B,C] :
           (rightangle(A,B,B,C) => perp(A,B,B,C)) )).
\end{lstlisting}

\item[\textbf{R7}] Two lines perpendicular to a third line are parallel to each
  other. %(Ludovic 27)
  This is rule D9 of the deductive database method~\cite{Chou2000}.

\begin{lstlisting}[language=GDDM]
fof(ruleD9,axiom,(![A,B,C,D,E,F] :
           (perp(A,B,E,F) & perp(C,D,E,F) => para(A,B,C,D) )).
\end{lstlisting}

\item[\textbf{R8}]  Two right triangles with two equal legs have equal hypotenuses.

\begin{lstlisting}[language=GDDM]
fof(ruleR8,axiom,(![A,B,C,D,E,F] :
            (rightangle(A,B,B,C) & rightangle(D,E,E,F) &
             cong(A,B,D,E) & cong(B,C,E,F) => cong(A,C,D,F)) )).
\end{lstlisting}

\item[\textbf{D58}] A rule, from the deductive database
  method~\cite{Chou2000}, needed for the proof of theorem~\ref{theo:problemOne}.

\begin{lstlisting}[language=GDDM]
  fof(ruleD58,axiom,(![A,B,C,P,Q,R] :
              (eqangle(A,B,B,C,P,Q,Q,R) & eqangle(A,C,B,C,P,R,Q,R) &
               ~coll(A,B,C)) => simtri(A,B,C,P,Q,R)) )).
\end{lstlisting}
\end{itemize}

\section{Some 7th Year Geometry Problems}
\label{sec:somethirdclassroomgeometryproblems}

In a step-wise learning, it is intended that there will be a
progressive proficiency in the use of hypothetical-deductive reasoning
and mathematical argumentation in primary and secondary school. It is
expected that, by the 7th year, students will be able to elaborate, with
some accuracy, small demonstrations~\cite{mec2013}.

To put this into practice the most appropriate domain is ``Geometry
and Measurement''~\cite{mec2013} or ``Geometry'' in the new curriculum
document~\cite{dge2021}.  In the seventh grade, as part of the study
of quadrilaterals, some properties of quadrilaterals and their
diagonals are studied. Using some concepts and results already learned,
equality of triangles, internal alternate angles,
corresponding angles, among others, students can and should be able to
demonstrate some of these properties, once they are armed with all the
necessary tools.  The following theorems illustrate some of these
properties, and are part of a larger set of results that can be
achieved.

It should also be noted that in this topic the use of technological
tools can be very valuable, in the drawing, in the execution of
rigorous constructions, in the visualisation of objects and their
properties, both by teachers and by students, when this use is done in
a thoughtful way.

\begin{figure}[htbp!]
  \centerline{\includegraphics[width=0.475\textwidth]{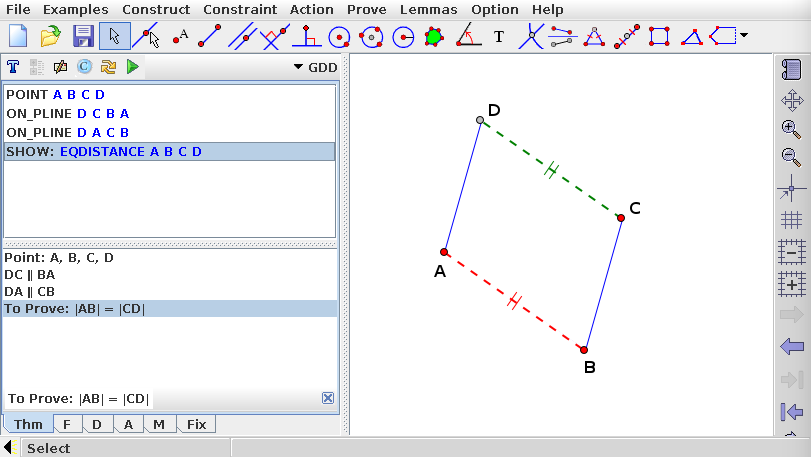}\quad\includegraphics[width=0.475\textwidth]{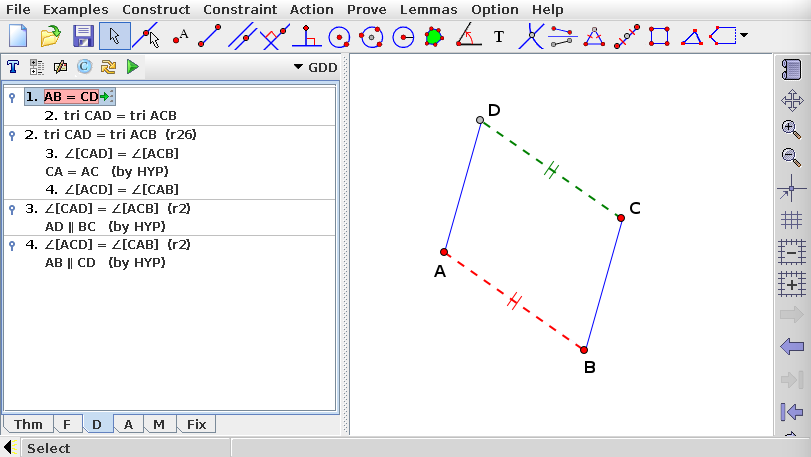}}
  \caption{Opposite Sides of a Parallelogram, in JGEx}
   \label{fig:problemOne}
\end{figure}

\begin{figure}[htbp!]
  \centerline{\includegraphics[width=0.475\textwidth]{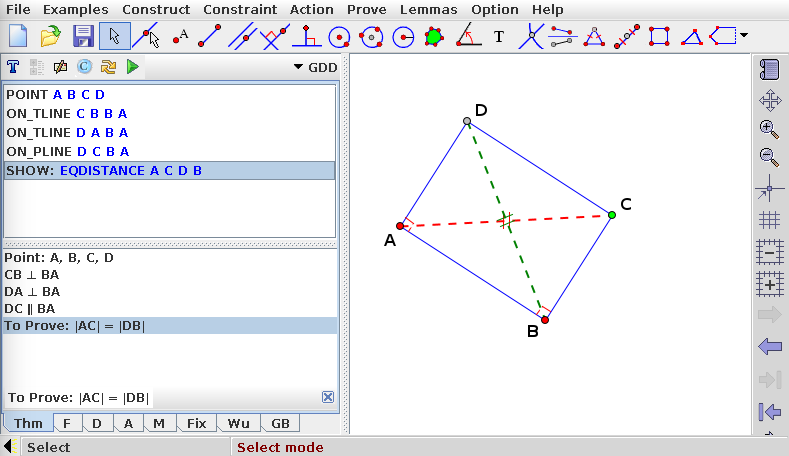}\quad\includegraphics[width=0.475\textwidth]{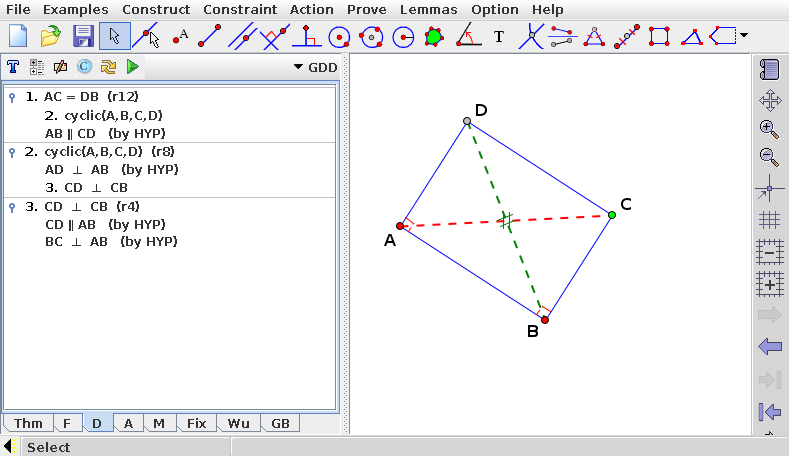}}
  \caption{Diagonals of a Rectangle, in JGEx}
  \label{fig:problemThree}
\end{figure}

The notation and terminology used in theorems and their proofs are
that defined in~\cite{mec2013} and currently used in Portuguese
textbooks. The geometric constructions, conjectures and proofs were
made with JGEx\footnote{The rule numbering used by the JGEx prover, e.g. ``(r26)'' are
  in no direct correspondence with the numbering of the rules described in
  this paper.} (see Fig.~\ref{fig:problemOne}--\ref{fig:problemThree}), the informal
demonstrations and the corresponding excerpts of formal demonstrations
were written by the authors, it is expected that those demonstrations
can be produced automatically by the new GATP being finished (see
Sec.~\ref{sec:conclusions}). Given that the proofs were done using the full-angle
rules~\cite{Chou1996b,Chou2000}, there are two different notions of
angles in use, the ``normal'' angle, defined by two semi-lines, are
used in the informal rules, the full-angles, defined by two lines, are
used in the formal rules. The conversion between the two definitions
are to be dealt by the process of rendering the formal proof in an
informal natural language form. Nevertheless this conversion it is not
without problems, for example, the known rule of triangles congruence,
side-angle-side (s.a.s.) rule, is not correct if full-angles are used~\cite{Chou2000}.

%\comentario{regras usadas: R1,R2,R3,R4,R5,R6, R12, R13, R14}

\bigskip
\begin{theorem}[Opposite Sides of a Parallelogram] \label{theo:problemOne} %  (Ludovic 63)
  If $[ABCD]$ is a parallelogram then the opposite sides are equal,
 i.e. $\overline{AB}=\overline{CD}$ and
  $\overline{AD}=\overline{BC}$ (see Figure~\ref{fig:problemOne}).
\end{theorem}

\begin{proof}

$[ABCD]$ is a parallelogram, by rule R1 (parallelogram definition),
the lines $AB$ and $CD$ are parallel and the lines $AD$ and $BC$ are
also parallel.

By rule R2, since the lines $AB$ and $CD$ are parallel, the angles
$BAC$ and $DCA$ are equal.

By rule R2, since the lines $AD$ and $BC$ are parallel, the angles
$ACB$ and $CAD$ are equal.

Since $BAC$ and $DCA$ are equal angles, $ACB$ and $CAD$ are equal
angles and, $\overline{AC}=\overline{CA}$, by rule R3 (a.s.a. criterion
of equality), triangles $[ABC]$ and $[CDA]$ are equal.

Finally, using rule R4, we have $\overline{AB}=\overline{CD}$ and
$\overline{BC}=\overline{DA}$.
\end{proof}

\begin{proof}[Excerpt of a Formal Proof done by a Rule-based GATP]
  As described in section~\ref{sec:gddm} a deductive database method
  prover works by forward chaining on the set of rules, starting on
  the conjecture and managing two sets of facts, ``new facts'' and
  ``old (already known) facts''.

  The following table tries to illustrate the way the prover would
  work in order to reach the intended conclusion. Many other facts
  would be generated by the application of the deductive database
  method rules, in its way to reach the fix-point. The presented chain
  of facts and rules would be obtained working backwards, from the
  conclusion to the conjecture, and it would be used to produce a
  natural-language, informal proof description. To avoid a longer  list of facts then necessary, the old facts list was written cumulatively
  from the start, i.e. only the new additions to the list are written.

  \medskip
\begin{lstlisting}[language=GDDM,frame=single]
include('geometryDeductiveDatabaseMethod.ax').
fof(theorem1,conjecture,(![A,B,C,D] :
                         parallelogram(A,B,C,D) => cong(A,B,C,D) & cong(A,D,B,C) )).
\end{lstlisting}

  \begin{longtable}{l|c|l|l}
    %%% Título
    \multicolumn{1}{c|}{\bf New Facts} & \multicolumn{1}{c|}{\bf
                                         Rules} &
                                                  \multicolumn{1}{c|}{\bf
                                                  Already Known Facts} &
                                                               \multicolumn{1}{c}{\bf
                                                               ndg.\footnote{ndg. ---
                                                               non-degenerated conditions.}}   \\ \hline
    %%% 1a linha
    \texttt{parallelogram(A,B,C,D)} & by hyp.  & \\ \hline
    \parbox{0.25\textwidth}{\tt
    para(A,B,C,D)\\
    para(A,D,B,C)} &   R1a,R1b  &\texttt{parallelogram(A,B,C,D)}
    \\ \hline
    %%% 2a linha
    \parbox{0.3\textwidth}{\tt
    eqangle(A,B,B,C,C,D,D,A)\\
    eqangle(A,C,B,C,C,A,D,A)} & D40 ($\times 2$) & \parbox{0.3\textwidth}{\tt
    % parallelogram(A,B,C,D)\\
    para(A,B,C,D)\\
    para(A,D,B,C)}  \\ \hline
  %%% 3a linha
    \parbox{0.3\textwidth}{\tt
    simtri(A,B,C,C,D,A)
    cong(B,D,B,D)\footnotemark} & D58 & \parbox{0.3\textwidth}{\tt
%                                 parallelogram(A,B,C,D)\\
%    para(A,B,C,D)\\
%    para(A,C,B,D) \\
    eqangle(A,B,B,C,C,D,D,A)\\
    eqangle(A,C,B,C,C,A,D,A)} & $\neg$\tt coll(A,B,C)  \\  \hline
  %%% 4a linha
    \parbox{0.3\textwidth}{\tt
    contri(A,B,C,C,D,A)} & D61 & \parbox{0.3\textwidth}{\tt
%                             parallelogram(A,B,C,D)\\
%    para(A,B,C,D)\\
%   para(A,C,B,D) \\
%    eqangle(A,B,B,C,C,D,D,A)\\
%    eqangle(A,C,B,C,C,A,D,A)\\
    simtri(A,B,C,C,D,A)} & $\neg$\tt coll(A,B,C)  \\  \hline
  %%% 5a linha
    \parbox{0.3\textwidth}{\tt
    cong(A,B,C,D) \\
    cong(D,A,B,C)} & R4a,R4b & \parbox{0.3\textwidth}{\tt
%                          parallelogram(A,B,C,D)\\
%    para(A,B,C,D)\\
  %  para(A,C,B,D) \\
  %  eqangle(A,B,B,C,C,D,D,A)\\
 %   eqangle(A,C,B,C,C,A,D,A)\\
 %   simtri(A,B,C,C,D,A) \\
    contri(A,B,C,C,D,A)
    } & $\neg$\tt coll(A,B,C)  \\
  \end{longtable}
  \footnotetext{Trivial fact that it is automatically added to apply the
    inference rule.}
\end{proof}

The formal proof, done by an actual implementation of the deductive
database method would have more steps, e.g. trivial steps, changing
the order of the letters to be able to apply the rules.  The
non-degenerated conditions (ndgs.) would be added automatically by the
prover, they must be added to the hypothesis to ensure that the
conjecture is true.  All this would lead to an informal
(rigorous) language rendering of the formal proof, hiding the trivial
steps (maybe with an explanation), explaining the need of the
non-degenerate conditions, and the conversion of the formal rules in a
«normal» secondary schools language, e.g. rule R3/D61 wrote as the
angle-side-angle (a.s.a.) rule for the congruence of triangles.

\bigskip
\begin{theorem}[Diagonals of a Rectangle]
  If $[ABCD]$ is a rectangle then its diagonals are equal, i.e.
  $\overline{AC}=\overline{BD}$ (see Figure~\ref{fig:problemThree}).
\end{theorem}

\begin{proof}
  If $[ABCD]$ is a rectangle by rule R5 the angles $BAD$,
  $CBA$, $DCB$ and $ADC$ are all right
  angles.

  By rule R6, the lines $AB$ and $AD$ are perpendicular, the lines
  $BC$ and $AB$ are perpendicular, the lines $BC$ and $CD$ are
  perpendicular and also the lines $AD$ and $CD$ are perpendicular.

  By rule R7, since $AB$ and $CD$ are both perpendicular to $AD$, $AB$
  and $CD$ are parallel, and since $AD$ and $BC$ are both
  perpendicular to $AB$, $ AD$ and $BC$ are parallel. ([ABCD] is a
  parallelogram, by rule R1)

  Repeating the steps of the proof of theorem 1 proof, we have
  $\overline{AD}=\overline{BC}$.

Since $\overline{AD}=\overline{BC}$, $BAD$ and
$CBA$ are both right angles and
$\overline{AB}=\overline{BA}$, by rule R8 (s.a.s. criterion of equality
of right triangles), the triangles $[ABD]$ and $[BAC]$ are equal.

Finally, using R4, we have $\overline{AC}=\overline{BD}$.
\end{proof}

  \medskip
\begin{lstlisting}[language=GDDM,frame=single]
include('geometryDeductiveDatabaseMethod.ax').
fof(theorem2,conjecture,(![A,B,C,D] : rectangle(A,B,C,D) => cong(A,C,B,D) )).
\end{lstlisting}

\begin{proof}[Excerpt of a Formal Proof done by a GDDM prover] {\ }\\

\begin{longtable}{l|c|l|l}
      %%% Título
      \multicolumn{1}{c|}{\bf New Facts} & \multicolumn{1}{c|}{\bf
        Rules} &
      \multicolumn{1}{c|}{\bf
        Already Known Facts} &
      \multicolumn{1}{c}{\bf
        ndg.}   \\ \hline
      %%% 1a linha
      \parbox{0.3\textwidth}{\tt
        rectangle(A,B,C,D)} & by hyp. & \\ \hline
      %%% 2a linha
       \parbox{0.3\textwidth}{\tt
         rightangle(D,A,A,B) \\
         rightangle(A,B,B,C)\\
         para(A,B,D,C)\\
         para(A,D,B,C)} & R5a,R5b,R5e,R5f & \parbox{0.3\textwidth}{\tt
         rectangle(A,B,C,D)}  \\ \hline
      %%% 3a linha
       \parbox{0.3\textwidth}{\tt
         rightangle(D,A,A,B) \\
         rightangle(A,B,B,C)\\
         parallelogram(A,B,C,D)} & R1 & \parbox{0.3\textwidth}{\tt
         para(A,B,D,C)\\
         para(A,D,B,C)}\\ \hline
      %%% 4a linha
       \parbox{0.3\textwidth}{\tt
         rightangle(D,A,A,B) \\
         rightangle(A,B,B,C)\\
         cong(A,B,C,D) \\
         cong(A,D,B,C)\\
         cong(A,B,A,B)\footnotemark} & Th. 1 & \parbox{0.3\textwidth}{\tt
         parallelogram(A,B,C,D)}\\ \hline
      %%% 5a linha
       \pagebreak \hline
       \parbox{0.3\textwidth}{\tt
         cong(A,B,C,D)\\
         cong(A,C,B,D)} & R8 & \parbox{0.3\textwidth}{\tt
         rightangle(D,A,A,B) \\
         rightangle(A,B,B,C)\\
         cong(A,D,B,C)\\
         cong(A,B,A,B)}
    \end{longtable}
    \footnotetext{Trivial fact that it is automatically added to apply
      the inference rule.}
\end{proof}

As can be seen in figure~\ref{fig:problemThree} the proof done by
JGEx uses a different approach. It uses the fact that the
points $A$, $B$, $C$ and $D$ are concyclic, and the diagonals of the
rectangle are diameters of the circle. This is not an approach
suitable for the intended school level, but, if JGEx is used
that would be the proof produced by it. An implementation of the
deductive database method should allow the selection of a given set of
rules, such possibility will be discussed in the final section (see
Section~\ref{sec:conclusions}).

\section{Lesson Plan}
\label{sec:lessonPlanoutline}

A lesson plan must reflect what will be proposed to students. Here an
outline of a lesson plan to address theorem~\ref{theo:problemOne} and its formal proof  by
students in a classroom is presented.  See appendix~\ref{sec:lessonPlan} for the lesson plan in full details.

The essential learning goals~\cite{dge2021} that this plan aims at are: explore, in a dynamic geometry environments (e.g. GeoGebra),  convex polygons with different numbers of sides; Formulate conjectures, generalisations and justifications, based on the identification of regularities common to the objects under study; establish conjectures in dynamic geometry environments and their exploration.

From those goals follow a set of tasks in which the student can conjecture about the various properties that are explored in the DGS.

\medskip

\textbf{Task 1} - Quadrilaterals, length of sides and measure of angles
\begin{itemize}
    \item Build a quadrilateral.
    \item Determine the length of the sides.
    \item Determine the measure of angles.
\end{itemize}

\smallskip

\textbf{Task 2} - Parallelogram
\begin{itemize}
    \item Build three parallelograms.
    \item Determine lengths of sides of those parallelograms
    \item What can you conjecture about the sides of a parallelogram?
\end{itemize}

\smallskip

\textbf{Task 3} - Prove the conjecture.
Follow the steps below:
\begin{itemize}
    \item Build a parallelogram $[ABCD]$
     \item Construct a diagonal of the parallelogram.
     \item Use the parallelism of the sides of the parallelogram and indicate the angles that have the same measure (mark these angles in your construction and indicate their measure)
     \item Remember the a.s.a. criterion of equality of triangles, in the figure which triangles can you conclude are equal?
     \item Mark in the figure with the same colour the segments that have the same length.
\end{itemize}

As already said, the goal is to explore geometric conjectures and their proofs, with the help of DGS and GATP computational tools.

In the final part of the lesson and before its conclusion, teacher will show to the students the full proof of the result using JGEx. Using the preview windows it is possible to follow the demonstration step by step, see the illustration of each step, and see how the formal rules  are being used. The teacher will have to adapt the formal language and the rules themselves into informal language closer to the language used by the students.

\section{Conclusions and Future Work}
\label{sec:conclusions}

This work aims to contribute  to the complex question of how to
get a $\approx\!12$-year-old student to understand and see the
connections between visualisation, construction and reasoning,  in
particular, the discursive processes to broaden the knowledge
processes, for demonstration and for interpretation.

The rules presented here (see Sec.~\ref{sec:thesetsofrules}) were
written in a rigorous mathematical (natural) language, appropriated to
be used in a classroom setting. Alongside those rules a corresponding
set of formal rules, appropriated to a rule base theorem prover were
also presented.  Using the set of rules found by Font~\cite{Font2021}
and the set of rules presented by Chou et al.~\cite{Chou2000} as a
starting point, our intention is to build a consistent set of rules
that can narrow the dissonance between the outcomes of the geometry
automated theorem provers and the normal practice of conjecturing and
proving in schools.

The use of a rule-based theorem prover being developed within our
research group~\cite{Baeta2022} will allow to implement the set of
rules which deem appropriated for the automation for one side and its use
with the intended target audience. There are many challenges ahead:
the implementation of an efficient prover; the possibility to separate
the set of rules from the inference mechanisms, i.e. the possibility
to adapt the set or rules to the task at hand without the need, each
time, to build a different GATP; the natural language rendering,
i.e. the transformation of the formal proof into a rigorous proof,
appropriated to be used in a classroom setting; the connection between
GeoGebra (a well-known, by the students and teachers, DGS) and the GATP,
i.e. the possibility of a visual connection between the construction,
the conjecture and the proof.

The lesson plan presented (see Sec.~\ref{sec:lessonPlan}) could be
applied at an actual classroom.  It is our intention to put into
practice an experience in a classroom, first through an experiment
with pre-service teachers and afterwords in an actual 7th year
classrooms.

\bibliographystyle{eptcs}
%\bibliography{geoTiles}
\bibliography{2022_TSQ_ThEdu22_AnIntroductionProofs7thYearWithRuleBasedTheoremProver}

\newcommand{\noopsort}[1]{}\newcommand{\singleletter}[1]{#1}
\begin{thebibliography}{10}
\providecommand{\bibitemdeclare}[2]{}
\providecommand{\surnamestart}{}
\providecommand{\surnameend}{}
\providecommand{\urlprefix}{Available at }
\providecommand{\url}[1]{\texttt{#1}}
\providecommand{\href}[2]{\texttt{#2}}
\providecommand{\urlalt}[2]{\href{#1}{#2}}
\providecommand{\doi}[1]{doi:\urlalt{http://dx.doi.org/#1}{#1}}
\providecommand{\bibinfo}[2]{#2}

\bibitemdeclare{article}{Baeta2022}
\bibitem{Baeta2022}
\bibinfo{author}{Nuno \surnamestart Baeta\surnameend} \& \bibinfo{author}{Pedro
  \surnamestart Quaresma\surnameend} (\bibinfo{year}{2022}):
  \emph{\bibinfo{title}{A Geometry Deductive Database Prover}}.
\newblock {\sl \bibinfo{journal}{Annals of Mathematics and Artificial
  Intelligence}}.
\newblock \bibinfo{note}{in press}.

\bibitemdeclare{incollection}{Chou2001}
\bibitem{Chou2001}
\bibinfo{author}{Shang-Ching \surnamestart Chou\surnameend} \&
  \bibinfo{author}{Xiao-Shan \surnamestart Gao\surnameend}
  (\bibinfo{year}{2001}): \emph{\bibinfo{title}{Automated Reasoning in
  Geometry}}.
\newblock In \bibinfo{editor}{John~Alan \surnamestart Robinson\surnameend} \&
  \bibinfo{editor}{Andrei \surnamestart Voronkov\surnameend}, editors: {\sl
  \bibinfo{booktitle}{Handbook of Automated Reasoning}},
  \bibinfo{publisher}{Elsevier Science Publishers B.V.}, pp.
  \bibinfo{pages}{707--749}, \doi{10.1016/B978-044450813-3/50013-8}.

\bibitemdeclare{article}{Chou1996b}
\bibitem{Chou1996b}
\bibinfo{author}{Shang-Ching \surnamestart Chou\surnameend},
  \bibinfo{author}{Xiao-Shan \surnamestart Gao\surnameend} \&
  \bibinfo{author}{Jing-Zhong \surnamestart Zhang\surnameend}
  (\bibinfo{year}{1996}): \emph{\bibinfo{title}{Automated generation of
  readable proofs with geometric invariants, {II}. Theorem Proving With
  Full-Angles}}.
\newblock {\sl \bibinfo{journal}{Journal of Automated Reasoning}}
  \bibinfo{volume}{17}(\bibinfo{number}{3}), pp. \bibinfo{pages}{349--370},
  \doi{10.1007/BF00283134}.

\bibitemdeclare{article}{Chou2000}
\bibitem{Chou2000}
\bibinfo{author}{Shang-Ching \surnamestart Chou\surnameend},
  \bibinfo{author}{Xiao-Shan \surnamestart Gao\surnameend} \&
  \bibinfo{author}{Jing-Zhong \surnamestart Zhang\surnameend}
  (\bibinfo{year}{2000}): \emph{\bibinfo{title}{A Deductive Database Approach
  to Automated Geometry Theorem Proving and Discovering}}.
\newblock {\sl \bibinfo{journal}{Journal of Automated Reasoning}}
  \bibinfo{volume}{25}(\bibinfo{number}{3}), p. \bibinfo{pages}{219–246},
  \doi{10.1023/A:1006171315513}.

\bibitemdeclare{book}{mec2013}
\bibitem{mec2013}
\bibinfo{author}{Ministério \surnamestart da~Educação~e
  Ciência~[MEC]\surnameend} (\bibinfo{year}{2013}):
  \emph{\bibinfo{title}{Programa de {Matemática} para o {Ensino} {Básico}}}.
\newblock \bibinfo{publisher}{Lisboa: Ministério da Educação e Ciência.}
\newblock
  \urlprefix\url{https://www.dge.mec.pt/sites/default/files/Basico/Metas/Matematica/programa_matematica_basico.pdf}.

\bibitemdeclare{incollection}{Duval}
\bibitem{Duval}
\bibinfo{author}{R~\surnamestart Duval\surnameend} (\bibinfo{year}{1998}):
  \emph{\bibinfo{title}{Geometry from a cognitive point of view}}.
\newblock \bibinfo{series}{Perspectives on the Teaching of Geometry for the
  21st century: an ICMI study}, \bibinfo{publisher}{Kluwer Academic}, pp.
  \bibinfo{pages}{37--51}.

\bibitemdeclare{book}{dge2018}
\bibitem{dge2018}
\bibinfo{author}{Direção-Geral \surnamestart da~Educação~[DGE]\surnameend}
  (\bibinfo{year}{2018}): \emph{\bibinfo{title}{{A}prendizagens essenciais:
  {A}rticulação com o perfil dos alunos - 7. º ano - 3.º ciclo do ensino
  básico – {M}atemática}}.
\newblock \bibinfo{publisher}{Lisboa: Direção-Geral da Educação.}
\newblock
  \urlprefix\url{http://www.dge.mec.pt/sites/default/files/Curriculo/Aprendizagens_Essenciais/3_ciclo/ae_mat_7.o_ano.pdf}.

\bibitemdeclare{book}{dge2021}
\bibitem{dge2021}
\bibinfo{author}{Direção-Geral \surnamestart da~Educação~[DGE]\surnameend}
  (\bibinfo{year}{2021}): \emph{\bibinfo{title}{{A}prendizagens essenciais:
  {A}rticulação com o perfil dos alunos - 7.º ano - 3.º ciclo do ensino
  básico – {M}atemática}}.
\newblock \bibinfo{publisher}{Lisboa: Direção-Geral da Educação.}
\newblock
  \urlprefix\url{https://www.dge.mec.pt/sites/default/files/Curriculo/Aprendizagens_Essenciais/3_ciclo/ae_mat_7.o_ano.pdf}.

\bibitemdeclare{phdthesis}{Font2021}
\bibitem{Font2021}
\bibinfo{author}{Ludovic \surnamestart Font\surnameend} (\bibinfo{year}{2021}):
  \emph{\bibinfo{title}{Génération automatique de preuves pour un logiciel
  tuteur en géométrie}}.
\newblock \bibinfo{type}{phdthesis}, \bibinfo{school}{Polytechnique Montréal}.
\newblock \urlprefix\url{https://publications.polymtl.ca/9090/}.

\bibitemdeclare{inproceedings}{Font2018}
\bibitem{Font2018}
\bibinfo{author}{Ludovic \surnamestart Font\surnameend},
  \bibinfo{author}{Philippe~R. \surnamestart Richard\surnameend} \&
  \bibinfo{author}{Michel \surnamestart Gagnon\surnameend}
  (\bibinfo{year}{2018}): \emph{\bibinfo{title}{Improving QED-Tutrix by
  Automating the Generation of Proofs}}.
\newblock In \bibinfo{editor}{Pedro \surnamestart Quaresma\surnameend} \&
  \bibinfo{editor}{Walther \surnamestart Neuper\surnameend}, editors: {\sl
  \bibinfo{booktitle}{Proceedings 6th International Workshop on Theorem proving
  components for Educational software, Gothenburg, Sweden, 6 Aug 2017}}, {\sl
  \bibinfo{series}{Electronic Proceedings in Theoretical Computer Science}}
  \bibinfo{volume}{267}, \bibinfo{publisher}{Open Publishing Association}, pp.
  \bibinfo{pages}{38--58}, \doi{10.4204/EPTCS.267.3}.

\bibitemdeclare{inproceedings}{Gagnon2017}
\bibitem{Gagnon2017}
\bibinfo{author}{M.~\surnamestart Gagnon\surnameend},
  \bibinfo{author}{N.~\surnamestart Leduc\surnameend}, \bibinfo{author}{P.R.
  \surnamestart Richard\surnameend} \& \bibinfo{author}{M.~\surnamestart
  Tessier-Baillargeon\surnameend} (\bibinfo{year}{2017}):
  \emph{\bibinfo{title}{QED-Tutrix: creating and expanding a problem database
  towards personalized problem itineraries for proof learning in geometry}}.
\newblock In: {\sl \bibinfo{booktitle}{Proceedings of the Tenth Congress of the
  European Society for Research in Mathematics Education (CERME10).}}

\bibitemdeclare{article}{Grothmann2016}
\bibitem{Grothmann2016}
\bibinfo{author}{René \surnamestart Grothmann\surnameend}
  (\bibinfo{year}{2016}): \emph{\bibinfo{title}{The Geometry Program C.a.R.}}
\newblock {\sl \bibinfo{journal}{International Journal of Computer Discovered
  Mathematics}} \bibinfo{volume}{1}(\bibinfo{number}{1}), pp.
  \bibinfo{pages}{45--61}.

\bibitemdeclare{article}{Hanna2020}
\bibitem{Hanna2020}
\bibinfo{author}{Gila \surnamestart Hanna\surnameend} (\bibinfo{year}{2020}):
  \emph{\bibinfo{title}{Mathematical Proof, Argumentation, and Reasoning}}, pp.
  \bibinfo{pages}{561--566}.
\newblock \doi{10.1007/978-3-030-15789-0\_102}.

\bibitemdeclare{mastersthesis}{Hohenwarter2002}
\bibitem{Hohenwarter2002}
\bibinfo{author}{M~\surnamestart Hohenwarter\surnameend}
  (\bibinfo{year}{2002}): \emph{\bibinfo{title}{GeoGebra - a software system
  for dynamic geometry and algebra in the plane}}.
\newblock Master's thesis, \bibinfo{school}{University of Salzburg},
  \bibinfo{address}{Austria}.

\bibitemdeclare{book}{Jackiw2001}
\bibitem{Jackiw2001}
\bibinfo{author}{N~\surnamestart Jackiw\surnameend} (\bibinfo{year}{2001}):
  \emph{\bibinfo{title}{The Geometer's Sketchpad v4.0}}.
\newblock \bibinfo{publisher}{Key Curriculum Press}.

\bibitemdeclare{incollection}{Janicic2006c}
\bibitem{Janicic2006c}
\bibinfo{author}{Predrag \surnamestart Jani\v{c}i\'c\surnameend}
  (\bibinfo{year}{2006}): \emph{\bibinfo{title}{{GCLC} --- {A} Tool for
  Constructive Euclidean Geometry and More Than That}}.
\newblock In \bibinfo{editor}{Andr\'es \surnamestart Iglesias\surnameend} \&
  \bibinfo{editor}{Nobuki \surnamestart Takayama\surnameend}, editors: {\sl
  \bibinfo{booktitle}{Mathematical Software - ICMS 2006}}, {\sl
  \bibinfo{series}{Lecture Notes in Computer Science}} \bibinfo{volume}{4151},
  \bibinfo{publisher}{Springer}, pp. \bibinfo{pages}{58--73},
  \doi{10.1007/11832225\_6}.

\bibitemdeclare{article}{Janicic2012a}
\bibitem{Janicic2012a}
\bibinfo{author}{Predrag \surnamestart Jani\v{c}i\'c\surnameend},
  \bibinfo{author}{Julien \surnamestart Narboux\surnameend} \&
  \bibinfo{author}{Pedro \surnamestart Quaresma\surnameend}
  (\bibinfo{year}{2012}): \emph{\bibinfo{title}{The {A}rea {M}ethod: a
  Recapitulation}}.
\newblock {\sl \bibinfo{journal}{Journal of Automated Reasoning}}
  \bibinfo{volume}{48}(\bibinfo{number}{4}), pp. \bibinfo{pages}{489--532},
  \doi{10.1007/s10817-010-9209-7}.

\bibitemdeclare{article}{Kovacs2018}
\bibitem{Kovacs2018}
\bibinfo{author}{Zoltan \surnamestart Kov\'acs\surnameend},
  \bibinfo{author}{Tomas \surnamestart Recio\surnameend} \&
  \bibinfo{author}{{Maria Pilar}. \surnamestart V\'elez\surnameend}
  (\bibinfo{year}{2018}): \emph{\bibinfo{title}{Using Automated Reasoning Tools
  in GeoGebra in the Teaching and Learning of Proving in Geometry}}.
\newblock {\sl \bibinfo{journal}{International Journal for Technology in
  Mathematics Education}} \bibinfo{volume}{25}(\bibinfo{number}{2}), pp.
  \bibinfo{pages}{33--50}, \doi{10.1564/tme\_v25.2.03}.

\bibitemdeclare{inbook}{Kovacs2022}
\bibitem{Kovacs2022}
\bibinfo{author}{Zoltan \surnamestart Kov\'acs\surnameend},
  \bibinfo{author}{Tomas \surnamestart Recio\surnameend} \&
  \bibinfo{author}{{Maria Pilar} \surnamestart V\'elez\surnameend}
  (\bibinfo{year}{2022}): \emph{\bibinfo{title}{Mathematics Education in the
  Age of Artificial Intelligence}}, chapter \bibinfo{chapter}{Automated
  Reasoning Tools with GeoGebra: What are they? What are they good for?}
\newblock \bibinfo{publisher}{Springer Nature},
  \doi{10.1007/978-3-030-86909-0\_2}.

\bibitemdeclare{article}{Laborde1990}
\bibitem{Laborde1990}
\bibinfo{author}{J.~M. \surnamestart Laborde\surnameend} \&
  \bibinfo{author}{R.~\surnamestart Str{\"a}sser\surnameend}
  (\bibinfo{year}{1990}): \emph{\bibinfo{title}{Cabri-G{\'e}om{\`e}tre: A
  microworld of geometry guided discovery learning}}.
\newblock {\sl \bibinfo{journal}{International reviews on mathematical
  education- Zentralblatt fuer didaktik der mathematik}}
  \bibinfo{volume}{90}(\bibinfo{number}{5}), pp. \bibinfo{pages}{171--177}.

\bibitemdeclare{article}{Marrades2000}
\bibitem{Marrades2000}
\bibinfo{author}{Ram{\'{o}}n \surnamestart Marrades\surnameend} \&
  \bibinfo{author}{{\'{A}}ngel \surnamestart Guti{\'{e}}rrez\surnameend}
  (\bibinfo{year}{2000}): \emph{\bibinfo{title}{Proofs produced by secondary
  school students learning geometry in a dynamic computer environment}}.
\newblock {\sl \bibinfo{journal}{Educational Studies in Mathematics}}
  \bibinfo{volume}{44}(\bibinfo{number}{1/2}), pp. \bibinfo{pages}{87--125},
  \doi{10.1023/A:1012785106627}.

\bibitemdeclare{inbook}{Quaresma2022e}
\bibitem{Quaresma2022e}
\bibinfo{author}{Pedro \surnamestart Quaresma\surnameend}
  (\bibinfo{year}{2022}): \emph{\bibinfo{title}{Evolution of Automated
  Deduction and Dynamic Constructions in Geometry}},
  chapter~\bibinfo{chapter}{1}, pp. \bibinfo{pages}{3--22}.
\newblock {\sl \bibinfo{series}{Mathematics Education in the Digital
  Era}}~\bibinfo{volume}{17}, \bibinfo{publisher}{Springer},
  \doi{10.1007/978-3-030-86909-0}.

\bibitemdeclare{inproceedings}{Quaresma2022a}
\bibitem{Quaresma2022a}
\bibinfo{author}{Pedro \surnamestart Quaresma\surnameend} \&
  \bibinfo{author}{Vanda \surnamestart Santos\surnameend}
  (\bibinfo{year}{2022}): \emph{\bibinfo{title}{Four Geometry Problems to
  Introduce Automated Deduction in Secondary Schools}}.
\newblock In: {\sl \bibinfo{booktitle}{Proceedings 10th International Workshop
  on Theorem Proving Components for Educational Software}}, {\sl
  \bibinfo{series}{Electronic Proceedings in Theoretical Computer Science}}
  \bibinfo{volume}{354}, \bibinfo{publisher}{Open Publishing Association}, pp.
  \bibinfo{pages}{27--42}, \doi{10.4204/eptcs.354.3}.

\bibitemdeclare{book}{Richter-Gebert1999}
\bibitem{Richter-Gebert1999}
\bibinfo{author}{J{\"u}rgen \surnamestart Richter-Gebert\surnameend} \&
  \bibinfo{author}{Ulrich \surnamestart Kortenkamp\surnameend}
  (\bibinfo{year}{1999}): \emph{\bibinfo{title}{The Interactive Geometry
  Software Cinderella}}.
\newblock \bibinfo{publisher}{Springer}.

\bibitemdeclare{article}{Santos2021}
\bibitem{Santos2021}
\bibinfo{author}{Vanda \surnamestart Santos\surnameend} \&
  \bibinfo{author}{Pedro \surnamestart Quaresma\surnameend}
  (\bibinfo{year}{2021}): \emph{\bibinfo{title}{Exploring Geometric Conjectures
  with the help of a Learning Environment - A Case Study with Pre-Service
  Teachers.}}
\newblock {\sl \bibinfo{journal}{The Electronic Journal of Mathematics and
  Technology}} \bibinfo{volume}{2}(\bibinfo{number}{1}).

\bibitemdeclare{incollection}{Ye2011}
\bibitem{Ye2011}
\bibinfo{author}{Zheng \surnamestart Ye\surnameend},
  \bibinfo{author}{Shang-Ching \surnamestart Chou\surnameend} \&
  \bibinfo{author}{Xiao-Shan \surnamestart Gao\surnameend}
  (\bibinfo{year}{2011}): \emph{\bibinfo{title}{An Introduction to {J}ava
  {G}eometry {E}xpert}}.
\newblock In \bibinfo{editor}{Thomas \surnamestart Sturm\surnameend} \&
  \bibinfo{editor}{Christoph \surnamestart Zengler\surnameend}, editors: {\sl
  \bibinfo{booktitle}{Automated Deduction in Geometry}}, {\sl
  \bibinfo{series}{Lecture Notes in Computer Science}} \bibinfo{volume}{6301},
  \bibinfo{publisher}{Springer Berlin Heidelberg}, pp.
  \bibinfo{pages}{189--195}, \doi{10.1007/978-3-642-21046-4\_10}.

\end{thebibliography}

% ---> geoTiles.bib
% <--- bibexport.bib
% bibexport 2022_TSQ_ThEdu22_AutomatedTheoremProving3rdGrade
% mv bibexport.bib 2022_TSQ_ThEdu22_AutomatedTheoremProving3rdGrade.bib

\appendix

\section{Lesson Plan}
\label{sec:lessonPlan}
\noindent
\textbf{Subject}: Geometry\\ %% ver avanço

\textbf{Subtopic}: Plane Figures\\

\textbf{Summary}: Properties of quadrilaterals: relationships between the lengths of the sides of a parallelogram. Using a dynamic geometry program to explore the tasks. Problem solving. Formal proof. \\

\textbf{Contents:} Quadrilaterals and interior angles.\\

\textbf{Background}
\begin{itemize}
    \item Parallelogram definition.
    \item Angles determined by two parallel lines intersected by a secant (alternate interior angles).
    \item Triangle equality criteria.
    \item Knowledge of using various tools including digital technology, namely specific computer programs (GeoGebra), in exploring properties of plane figures.
\end{itemize}

\textbf{Essential Learning} \\
Explore, in dynamic geometry environments, e.g. GeoGebra, convex polygons with different numbers of sides.\\
Formulate conjectures, generalisations and justifications, based on the identification of regularities common to the objects under study.\\
Establish conjectures in dynamic geometry environments, e.g. GeoGebra, and their exploration.\\

\textbf{Methodology/Strategy}\\
Dialogue exposition. Exploration of dynamic geometry environments. Generate individual and group work opportunities.\\

\textbf{Teaching resources/materials}
\begin{itemize}
    \item Computer
    \item Dynamic geometry system
    \item Projector
    \item PowerPoint to support tasks to be proposed to students
    \item Activity sheet ``Let's explore the parallelogram''
\end{itemize}

\textbf{Classroom organisation}\\
The organisation of the classroom space is in a computer room.\\

\textbf{Class Description - Class Strategies and Development}
\begin{itemize}
    \item Presentation of the task and the way of working in class (10 minutes)
\end{itemize}
After the students enter the classroom, the lesson is opened and the summary is written on the board. Students record the summary in their daily notebook. Then the teacher will propose to them the task ``Let's explore the parallelogram''. The teacher will invite students to form pairs and choose a computer to work on. The teacher explains to the students that they will solve tasks, in a previously available form, in a group on the computer.\\

The dynamic geometry environment will be explored by performing the following tasks (total of 50 minutes):\\

\textbf{Task 1} - Quadrilaterals, length of sides and measure of angles (10 minutes).
\begin{itemize}
    \item Build a quadrilateral.
    \item Determine the length of the sides.
    \item Determine the measure of angles.
\end{itemize}

\textbf{Task 2} - Parallelogram (10 minutes).
\begin{itemize}
    \item Build three parallelograms.
    \item Determine lengths of sides of those parallelograms
    \item What can you conjecture about the sides of a parallelogram?
\end{itemize}

\textbf{Task 3} - Prove the conjecture (30 minutes).
Follow the steps below:
\begin{itemize}
    \item Build a parallelogram $[ABCD]$.
     \item Construct a diagonal of the parallelogram.
     \item Use the parallelism of the sides of the parallelogram and
       indicate the angles that have the same measure (mark those
       angles in your construction and indicate their measure).
     \item Remember the a.s.a. criterion for equality of triangles, in the figure which triangles can you conclude are equal?
     \item Mark with the same colour, in the figure, the segments that have the same length.
\end{itemize}

Complete the following sentence:

\hspace{35pt}``In a parallelogram,   \ldots sides are \ldots''

\begin{itemize}
    \item Task discussion (10 minutes)
\end{itemize}

\begin{itemize}
    \item Synthesis and formal proof (20 minutes)
\end{itemize}

A synthesis will be carried out in the last part of carrying out the tasks. In fact, this will be done at the same time as the discussion.

In the final part of the lesson and before its conclusion, the teacher will show to the students the full proof of the result using JGEx. Through the preview windows it is possible to follow the demonstration step by step, see the illustration of each step, and see how the formal rules  are being used. The teacher will have to adapt the formal language and the rules themselves into informal language closer to the language used by the students.\\

\textbf{Time schedule}

\begin{itemize}
 \item Task introduction: 10 minutes
 \item Autonomous task work: 50 minutes
 \item Discussion: 10 minutes
 \item Synthesis and formal proof: 20 minutes
\end{itemize}

\end{document}